\documentclass[sigconf, nonacm]{acmart}

\setcopyright{none}
\renewcommand\footnotetextcopyrightpermission[1]{}
\usepackage{colortbl}
\usepackage{graphicx} 
\usepackage{subcaption} 
\usepackage{adjustbox}
\usepackage{float}
\usepackage{multirow}

\begin{document}

\title{GroMo: Plant Growth Modeling with Multiview Images}

 
\author{Ruchi Bhatt}
\email{ruchi.21csz0007@iitrpr.ac.in}
\affiliation{%
  \institution{Indian Institute of Technology Ropar}
  \city{}
  \country{India}
}

\author{Shreya Bansal}
\email{shreya.22csz0010@iitrpr.ac.in}
\affiliation{%
  \institution{Indian Institute of Technology Ropar}
  \city{}
  \country{India}
}

\author{Amanpreet Chander}
\email{2018bmz0002@iitrpr.ac.in}
\affiliation{%
  \institution{Indian Institute of Technology Ropar,}
  \city{}
  \country{India}
}

\author{Rupinder Kaur}
\email{rupinder.23csz0008@iitrpr.ac.in}
\affiliation{%
  \institution{Indian Institute of Technology Ropar,}
  \city{}
  \country{India}
}

\author{Malya Singh}
\email{malya.22csz0015@iitrpr.ac.in}
\affiliation{%
  \institution{Indian Institute of Technology Ropar,}
  \city{}
  \country{India}
}

\author{Mohan Kankanhalli}
\email{mohan@comp.nus.edu.sg}
\affiliation{%
  \institution{National University of Singapore,}
  \city{}
  \country{Singapore}
}

\author{Abdulmotaleb El Saddik}
\email{elsaddik@uottawa.ca}
\affiliation{%
  \institution{University of Ottawa,}
  \city{}
  \country{Canada}
}


\author{Mukesh Kumar Saini}
\email{mukesh@iitrpr.ac.in}
\affiliation{%
  \institution{Indian Institute of Technology Ropar}
  \city{}
  \country{India}
}



\begin{abstract}

Understanding plant growth dynamics is essential for applications in agriculture and plant phenotyping. We present the Growth Modelling (GroMo) challenge, which is designed for two primary tasks: (1) plant age prediction and (2) leaf count estimation, both essential for crop monitoring and precision agriculture. For this challenge, we introduce GroMo25, a dataset with images of four crops: radish, okra, wheat, and mustard. Each crop consists of multiple plants (p1, p2, ..., pn) captured over different days (d1, d2, ..., dm) and categorized into five levels (L1, L2, L3, L4, L5). Each plant is captured from 24 different angles with a 15-degree gap between images. Participants are required to perform both tasks for all four crops with these multiview images. We proposed a Multiview Vision Transformer (MVVT) model for the GroMo challenge and evaluated the crop-wise performance on GroMo25. MVVT reports an average MAE of 7.74 for age prediction and an MAE of 5.52 for leaf count. The GroMo Challenge aims to advance plant phenotyping research by encouraging innovative solutions for tracking and predicting plant growth. The github repository is publicly available at \url{https://github.com/mriglab/GroMo-Plant-Growth-Modeling-with-Multiview-Images}.

\end{abstract}







\maketitle

\begin{figure*}[th] 
    \centering
    \begin{subcaptionbox}{}{\includegraphics[width=0.19\textwidth, alt= {Mustard}]{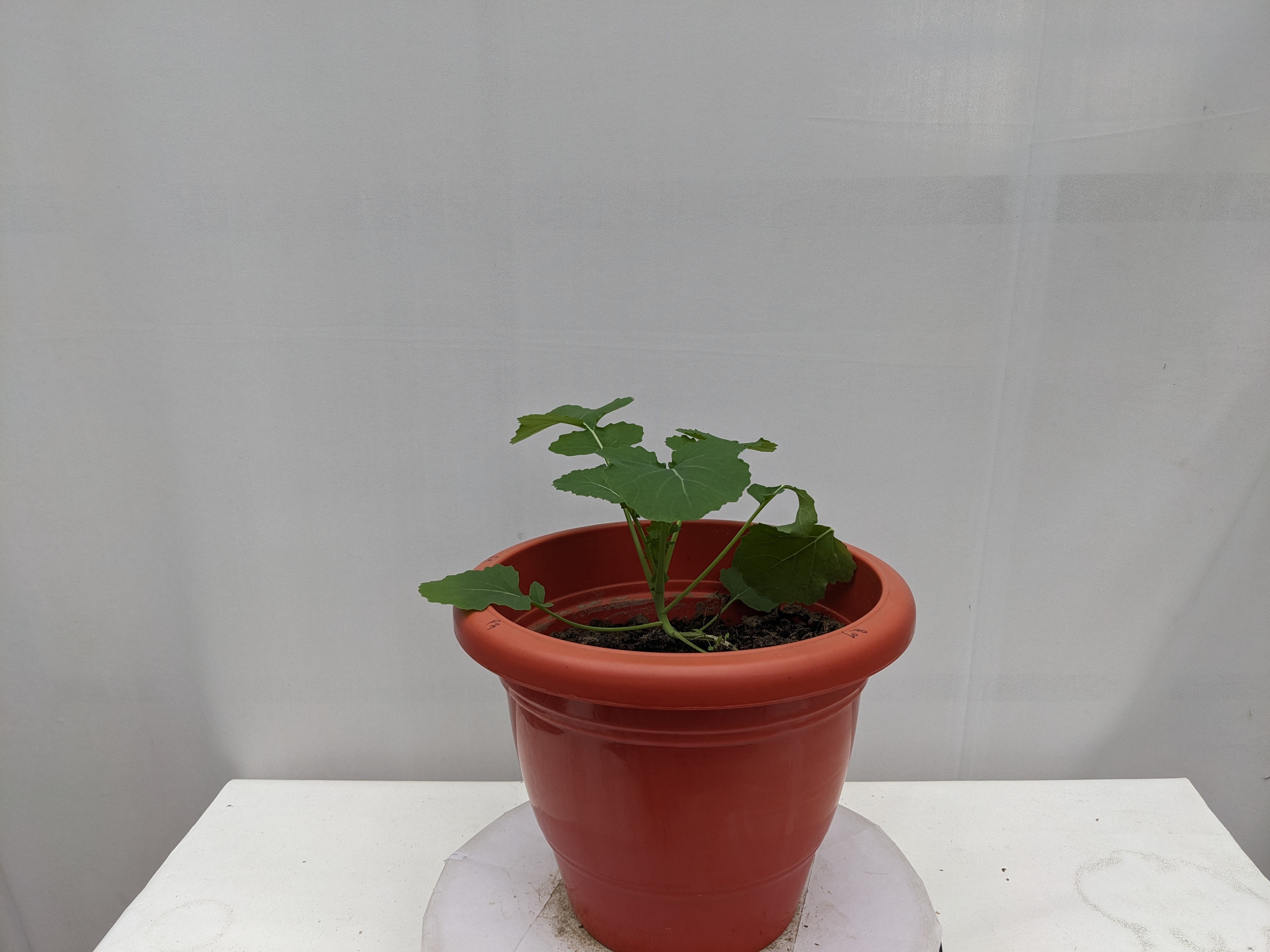}}\end{subcaptionbox}
    \begin{subcaptionbox}{}{\includegraphics[width=0.19\textwidth, alt={Radish}]{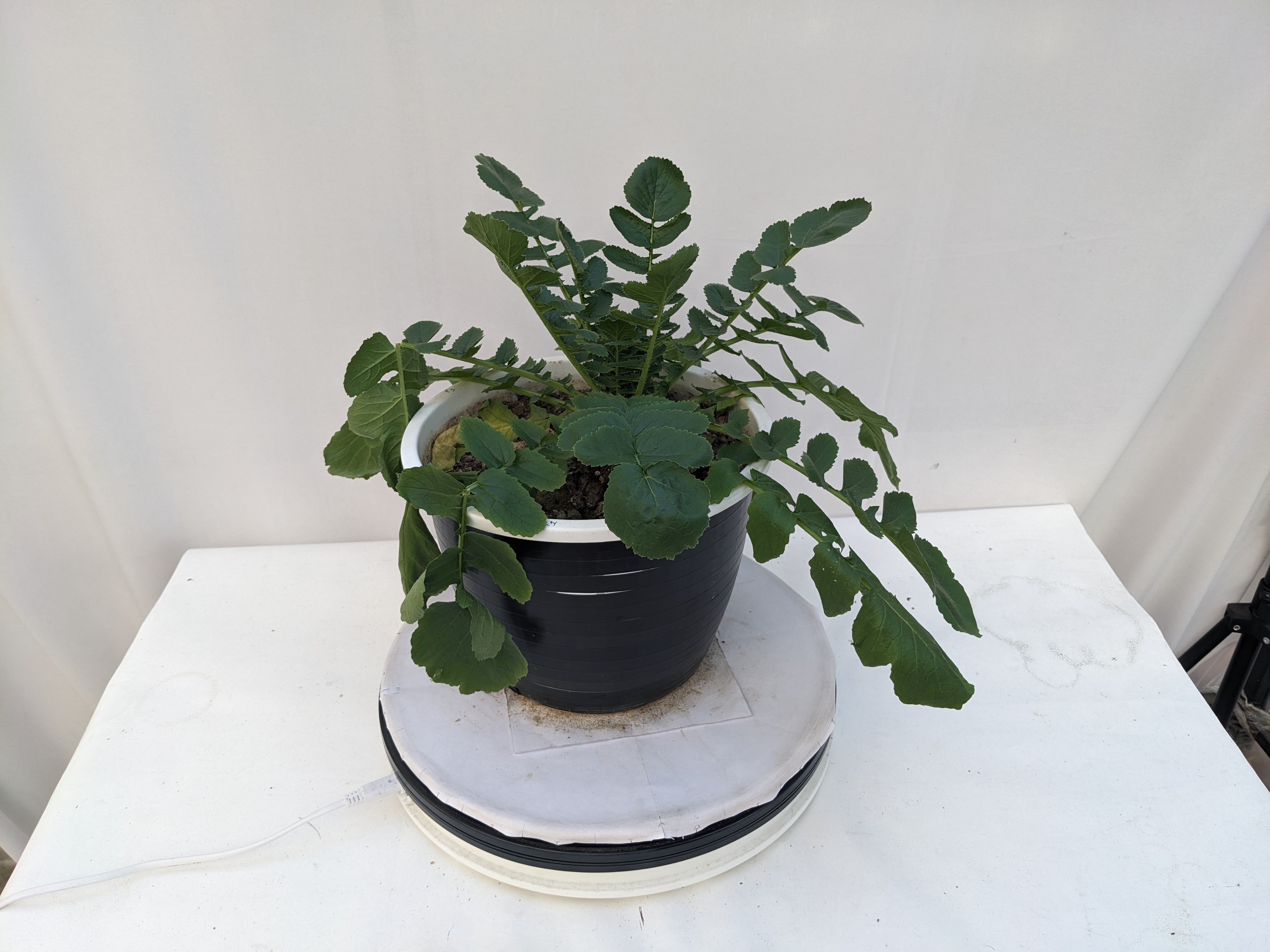}}\end{subcaptionbox}
    \begin{subcaptionbox}{}{\includegraphics[width=0.19\textwidth, alt={Okra}]{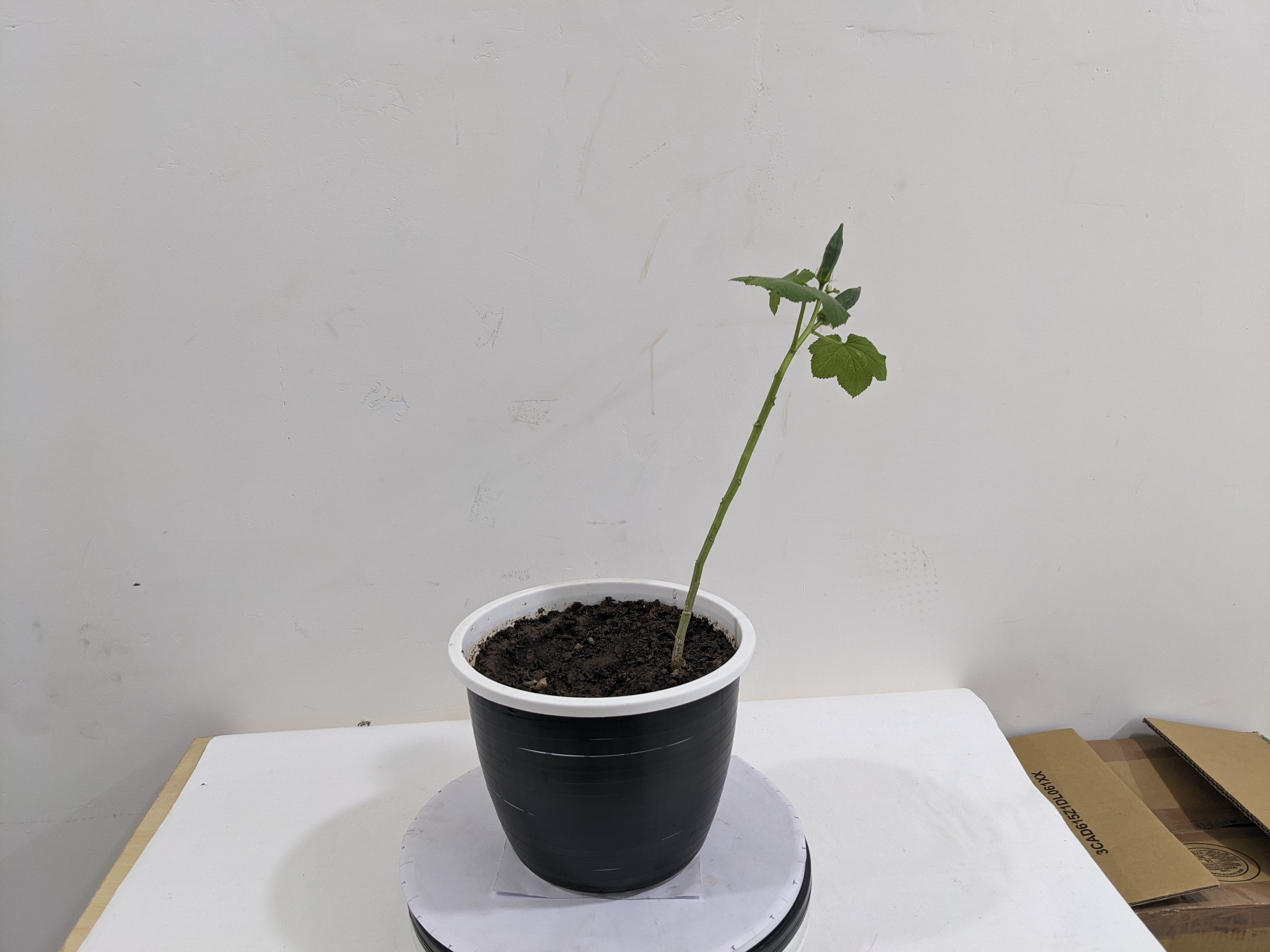}}\end{subcaptionbox}
    \begin{subcaptionbox}{}{\includegraphics[width=0.19\textwidth, alt={Wheat}]{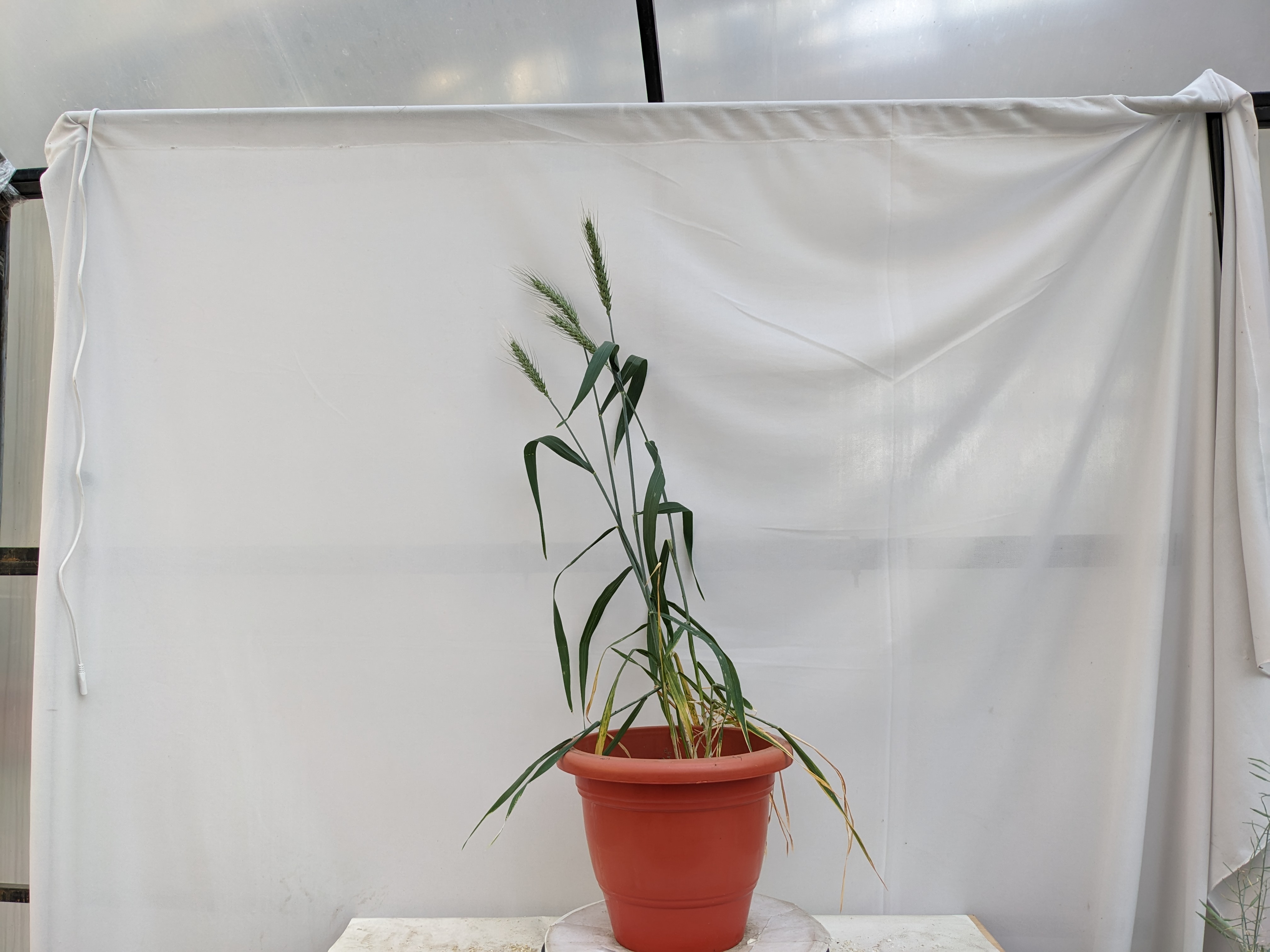}}\end{subcaptionbox}
    \caption{GrowMo25 sample of four crops: (a) Mustard (b) Radish (c) Okra and (d) Wheat.}
    \label{fig: Crops}
 \Description{GrowMo25 sample of four crops: (a) Mustard (b) Radish (c) Okra and (d) Wheat.}
\end{figure*}

\section*{Introduction }
Plant growth monitoring is crucial for plant breeding, precision agriculture, and yield estimation. Growth can be assessed by tracking key plant organs such as leaves \cite{b19}, flowers \cite{b20}, stems \cite{b21}, and fruits \cite{b22}, as changes in these phenotypes serve as indicators of development over time. Leaf counting, in particular, is a fundamental task in plant growth estimation and analysis \cite{b25}. We introduce a multi-view time-series plant dataset along with two challenging tasks: plant age estimation and leaf counting. While existing datasets for leaf counting primarily focus on top-view images, our dataset provides a massively multi-view perspective to address occlusion and improve growth estimation accuracy. In addition, the time-series data allow researchers to estimate the plant's age and predict the growth progression. We believe this dataset will open new possibilities for plant phenotyping using multimedia tools. 

\section*{Related work}
Multiple studies have been conducted in the domain of growth estimation via leaf counting. Farjon et al. \cite{b26} used two network architectures, one utilizing direct regression and the other integrating regression with detection. Dobrescu et al. \cite{b25} investigated the performance of deep learning models in plant phenotyping, specifically in the leaf counting regression task. Their study found that the model focuses primarily on the plant rather than the background, with leaf edges being the most significant features for prediction \cite{b27}. They used the VGG-16 deep learning model on the CVPPP 2017 Leaf Counting Challenge dataset. Buzzy et al. \cite{b32} trained a Tiny-YOLOv3 model for accurate localization and leaf counting operations. The model training was done using Arabidopsis plant images captured using a Canon Rebel XS camera. Shubra and Ian \cite{b33} investigated the problem of counting rosette leaves from RGB images. They used a deconvolutional network for initial segmentation and a convolutional network for leaf counting. 
Bhagat et al. \cite{b34} performed segmentation and leaf counting using Eff-UNet++, an encoder-decoder-based architecture. They evaluated their model on the KOMATSUNA dataset, the Multi-Modality Plant Imagery Dataset (MSU-PID), and the Computer Vision for Plant Phenotyping dataset (CVPPP). Fan et al. \cite{b35} proposed a two-stream deep learning framework with a spatial pyramid structure for segmentation and leaf counting. Deb et al., in 2024, proposed a convolution neural network-based leaf counting architecture named LC-Net. They utilized the SegNet model to obtain segmented leaf parts \cite{b36}. Table \ref{tab:plant_phenotyping} summarises previous related challenges and corresponding datasets. Most of the datasets above have a fixed camera view, leading to occlusion issues that affect leaf counting and growth estimation. Also, we could not find any work that estimates the age of a plant in a given image. The proposed dataset and challenge offer the opportunity to use multimedia techniques for enhanced plant growth modeling.    

\begin{table}[ht!]
\centering
\renewcommand{\arraystretch}{1.2} 
\begin{tabular}{p{2.2cm} p{2.6cm} p{2.4cm}}
\hline
\textbf{Model Used} & \textbf{Dataset} & \textbf{Challenge} \\ \hline
\begin{tabular}[c]{@{}l@{}}Fusing Network \\ Components \cite{b25} \end{tabular} & 
\begin{tabular}[c]{@{}l@{}}Dobrescu et al., \\ 2017 \end{tabular} & 
\begin{tabular}[c]{@{}l@{}}Counting Challenge \\ (LCC) \end{tabular} \\ \hline

\begin{tabular}[c]{@{}l@{}}VGG-16 \\ network \cite{b27} \end{tabular} & 
\begin{tabular}[c]{@{}l@{}}CVPPP 2017 dataset \\ (Plant Phenotyping) \end{tabular} & 
\begin{tabular}[c]{@{}l@{}}CVPPP Leaf \\ Counting Challenge \end{tabular} \\ \hline

\begin{tabular}[c]{@{}l@{}}Tiny-YOLOv3 \\ network \cite{b31} \end{tabular} & 
\begin{tabular}[c]{@{}l@{}}Arabidopsis \\ thaliana \end{tabular} & 
- \\ \hline

\begin{tabular}[c]{@{}l@{}}Deconvolutional \\ network \cite{b33} \end{tabular} & 
\begin{tabular}[c]{@{}l@{}}CVPPP-2017 \\ dataset \end{tabular} & 
\begin{tabular}[c]{@{}l@{}}CVPPP Leaf \\ Counting Challenge \end{tabular} \\ \hline

\begin{tabular}[c]{@{}l@{}}Eff-UNet++ \cite{b34} \end{tabular} & 
\begin{tabular}[c]{@{}l@{}}KOMATSUNA, \\ MSU-PID, \\ CVPPP-2017 \end{tabular} & 
\begin{tabular}[c]{@{}l@{}}CVPPP Leaf \\ Counting Challenge \end{tabular} \\ \hline

\begin{tabular}[c]{@{}l@{}}Two-stream CNN \\ \cite{b35} \end{tabular} & 
\begin{tabular}[c]{@{}l@{}}CVPPP 2017 dataset \end{tabular} & 
\begin{tabular}[c]{@{}l@{}}CVPPP Leaf \\ Counting Challenge \end{tabular} \\ \hline

\begin{tabular}[c]{@{}l@{}}LC-Net \cite{b36} \end{tabular} & 
\begin{tabular}[c]{@{}l@{}}CVPPP, \\ KOMATSUNA \\ datasets \end{tabular} & 
\begin{tabular}[c]{@{}l@{}}CVPPP Leaf \\ Counting Challenge \end{tabular} \\ \hline
\end{tabular}
\caption{Summary of models, datasets, and challenges for plant phenotyping tasks.}
\label{tab:plant_phenotyping}
\end{table}

\section*{Dataset }
The GroMo25 dataset consists of images collected from four crops: wheat, mustard, radish, and okra, each with multiple plant instances as shown in Figure \ref{fig: Crops}. Collected in a controlled environment, the dataset ensures consistency across observations. Each potted plant was placed on a rotator device to capture comprehensive multi-view representations, allowing images to be taken from multiple angles at different observational levels. The dataset spans the full growth duration of each crop, capturing detailed visual changes over time.

\subsection*{Dataset Statistics}
Table \ref{tab:dataset_summary} provides a structured overview of the dataset, summarizing key attributes for each crop.

\begin{table}[ht!]
    \centering
    
    \begin{tabular}{@{}lccccc@{}}
        \toprule
        \textbf{Crop} & \textbf{Plants} & \textbf{Max Days} & \textbf{Levels} & \textbf{Angles} \\
        \midrule
        Wheat   & 4 & 118 & 5 & 0\textdegree-360\textdegree (step 15\textdegree) \\
        Mustard & 4 & 50  & 5 & 0\textdegree-360\textdegree (step 15\textdegree) \\
        Radish  & 5 & 59  & 5 & 0\textdegree-360\textdegree (step 15\textdegree) \\
        Okra    & 2 & 86  & 5 & 0\textdegree-360\textdegree (step 15\textdegree) \\
        \bottomrule
    \end{tabular}
    
    \caption{Dataset Summary}
    \label{tab:dataset_summary}
\end{table}

\begin{figure*}
    \centering
    \includegraphics[width=1\linewidth, alt={day wise}]{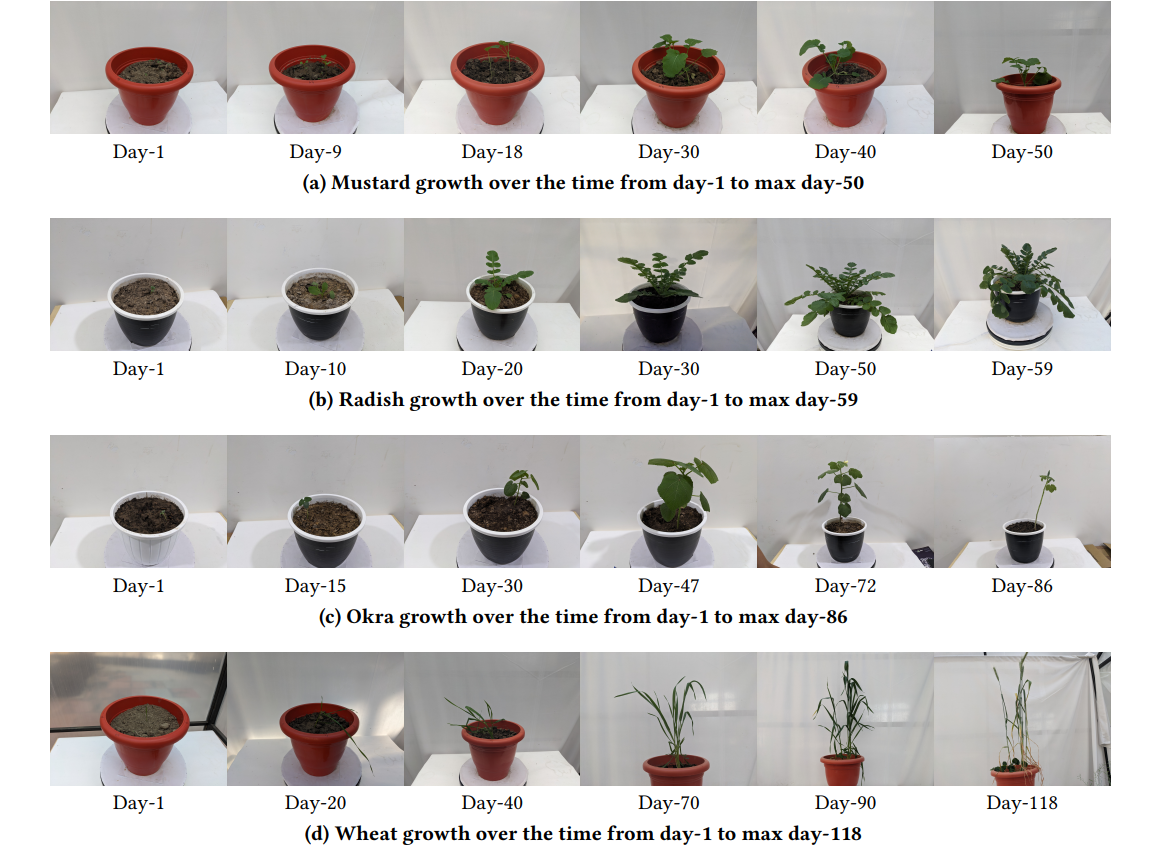}
   \caption{ Sample images of different crops at different intervals of growth (d1 to max\_day).}
\Description{ Sample images of different crops at different intervals of growth (d1 to max\_day).}
    \label{fig:days}
\end{figure*}
\begin{figure*}
    \centering
    \includegraphics[width=1\linewidth, alt={level wise}]{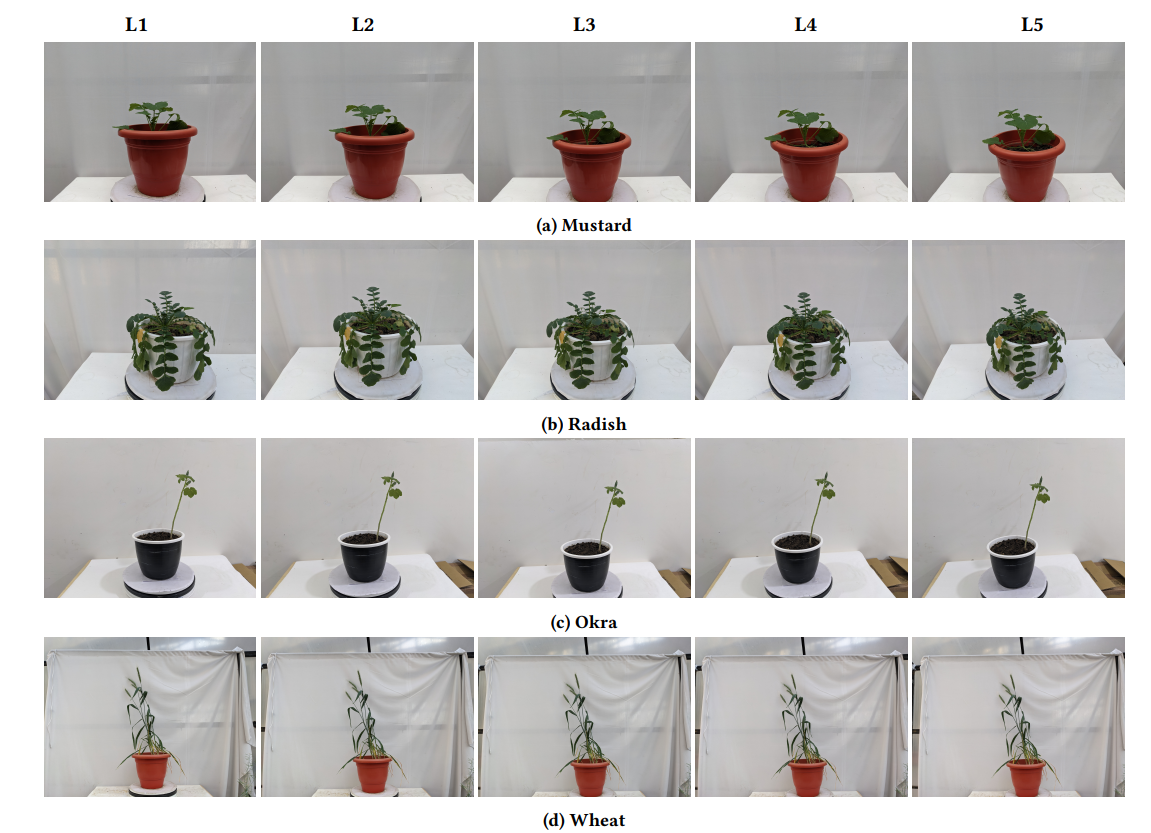}
    \caption{Sample images of different crops at five different levels (L1 to L5). }
      \Description{Sample images of different crops at five different levels (L1 to L5).}
    \label{fig:levels}
\end{figure*}
\begin{figure*}
    \centering
    \includegraphics[width=1\linewidth, alt={Anfle wise}]{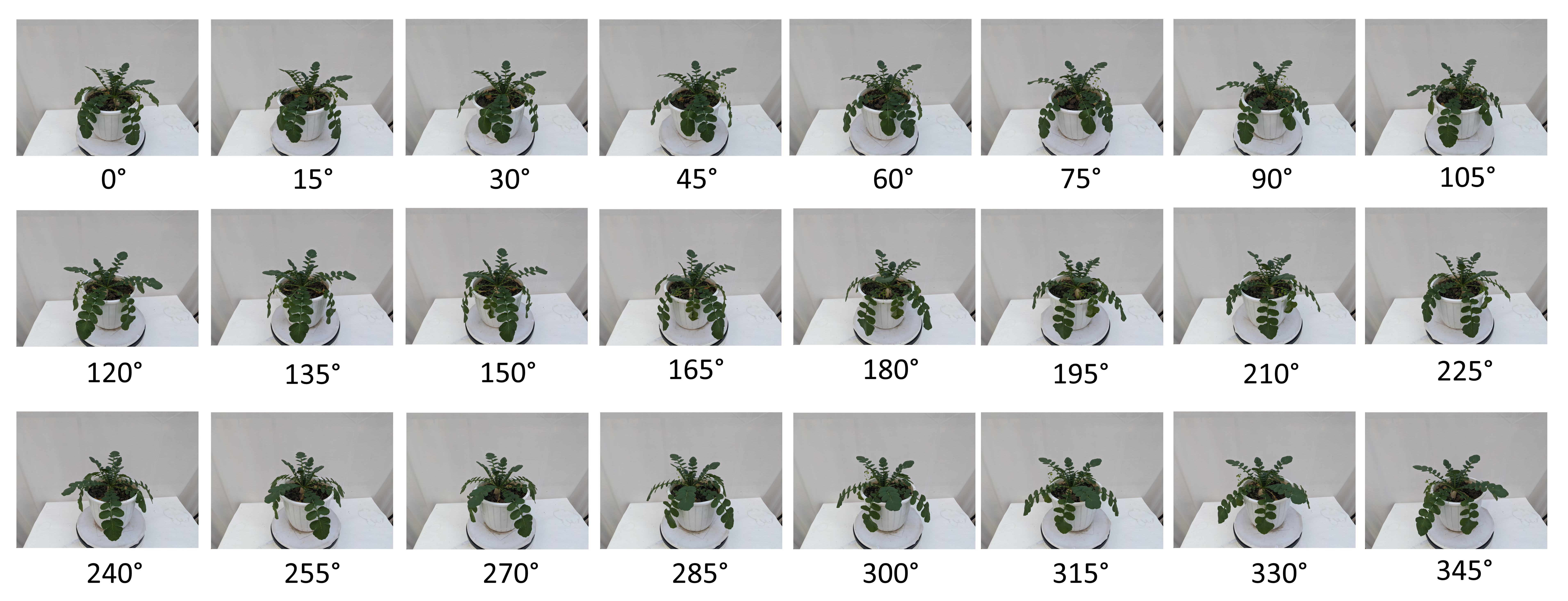}
    \caption{Each plant of radish is captured from 24 different angles with a 15-degree gap between images. }
      \Description{Each plant of radish is captured from 24 different angles with a 15-degree gap between images.}
    \label{fig:angle_shift}
\end{figure*}

\begin{itemize}
    \item \textbf{Plants:} The number of plant instances per crop varies, ensuring diversity in the dataset. Wheat and mustard have four plant instances each, radish has five, and okra has two. This variation captures different growth patterns and structural differences within the same crop category.

    \item \textbf{Max Days:} The dataset spans the full growth cycle of each crop, with the maximum number of observation days depending on the crop type. Wheat has the longest observation period (118 days), followed by okra (86 days), radish (59 days), and mustard (50 days). See Figure \ref{fig:days}.

    \item \textbf{Levels:} To ensure a detailed analysis of plant growth, images were captured at five different observational levels (L1–L5). These levels represent different heights, allowing for a comprehensive view of the plant structure from the base to the top. See Figure \ref{fig:levels}.

    \item \textbf{Angles:} Multi-view image capture was achieved by rotating each plant through a full 360° using a rotator device. Images were taken at 15° intervals, resulting in 24 different perspectives per plant for each observation. This setup ensures that plant morphology is well-documented from all possible viewing angles. See Figure \ref{fig:angle_shift}.
\end{itemize}
The structured approach in data collection ensures that the dataset provides extensive coverage of plant growth patterns, making it suitable for age prediction and leaf-counting tasks.

\section*{Tasks and Challenges}
\begin{figure*}
    \centering
    \includegraphics[width=1\linewidth, alt={Diagram of the Multi-View Vision Transformer model}]{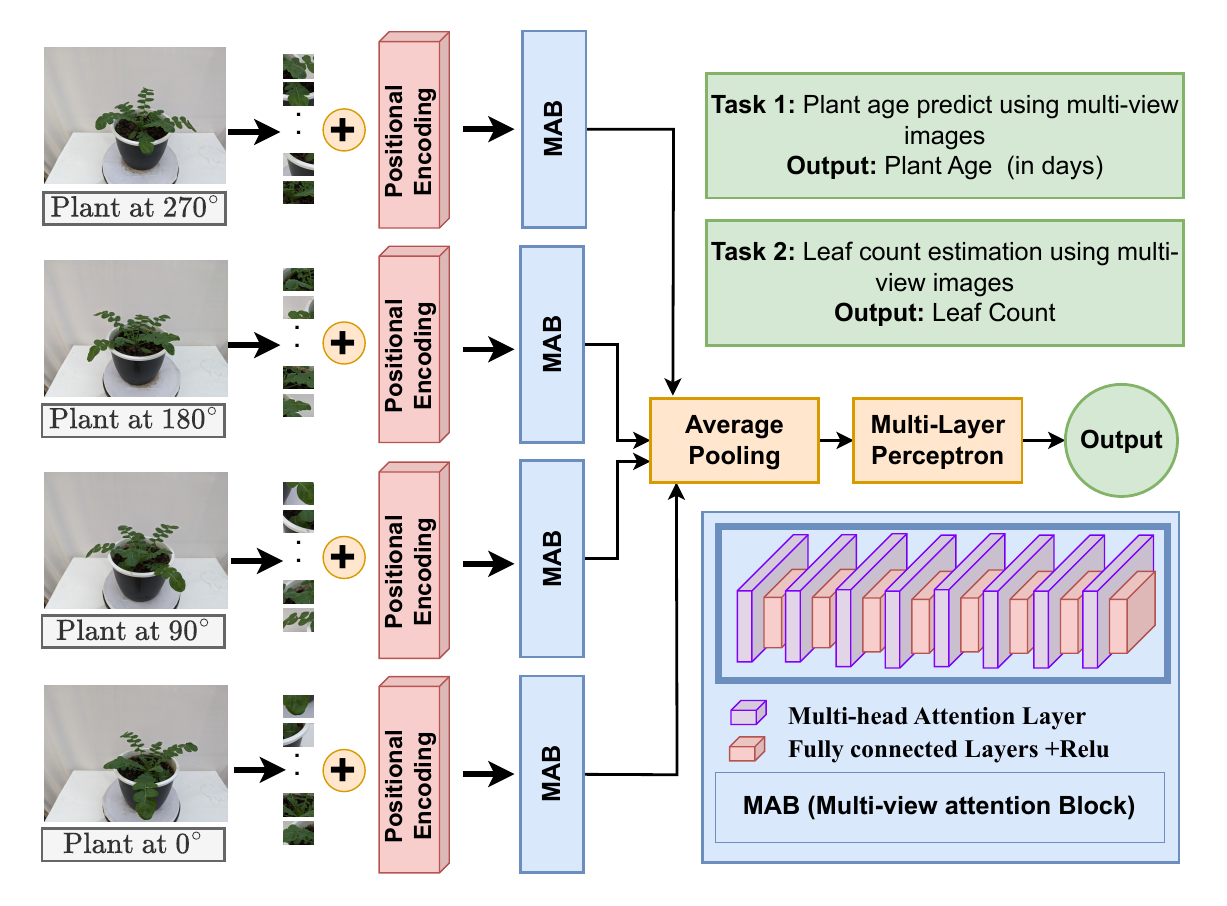}
    \caption{The figure shows the Multi-View Vision Transformer (MVVT)  model which processes multi-view images by first embedding each image into patches, incorporating positional encodings to retain spatial relationships, and applying a Multi-view Attention Block to capture dependencies across views. These representations are passed through a transformer encoder to capture complex relationships, followed by mean pooling and an MLP head for the final task. }
      \Description{ The figure shows the Multi-View Vision Transformer (MVVT)  model which processes multi-view images by first embedding each image into patches, incorporating positional encodings to retain spatial relationships, and applying a Multi-view Attention Block to capture dependencies across views. These representations are passed through a transformer encoder to capture complex relationships, followed by mean pooling and an MLP head for the final task.}
    \label{fig:model}
\end{figure*}

The GroMo Challenge requires participants to develop models that effectively combine multiview images to predict plant age and leaf count. Using up to 24 images taken from different angles and height levels, participants must fuse this information to improve plant growth modeling. The challenge is based on a collected dataset and consists of two tasks: predicting plant age in days and estimating the number of leaves. For both tasks, participants must build models using data from different crops. The performance of their models will be evaluated using Root Mean Square Error (RMSE), and the final ranking will be determined by averaging the results across all crops for each task. This challenge aims to advance techniques for analyzing plant development through diverse visual perspectives. The two tasks are mentioned below in brief:

\begin{enumerate}
    \item {\textbf{Task 1 - Plant Age Prediction:} \\Participants must develop a model that predicts the age of a plant in days using multiple views of the same plant. The dataset for each crop should be used separately for training and validation. The dataset consists of images captured at five different height levels, and participants must incorporate all five levels in their model. The participants can vary the number of images per level to cover a 360° view. The accuracy of predictions will be assessed using RMSE, with results reported separately for each crop. The final evaluation for this task will be based on the average RMSE across all crops.}

    \item {\textbf{Task 2 - Leaf Count Estimation:} \\Participants must build a model that counts the number of leaves on a plant using multiple views of the same plant. The dataset for each crop should be used separately for training and validation. The dataset consists of images captured at five different height levels, and participants must incorporate all five levels in their model. The participants can vary the number of images per level to cover a 360° view. The leaf count estimation will be assessed using RMSE, with results reported separately for each crop. The final evaluation for this task will be based on the average RMSE across all crops.}

\end{enumerate}




\section*{Methodology}
We propose a Multi-View Vision Transformer (MVVT) model, an extension of the Vision Transformer (ViT)\cite{b37}, designed to process multi-view image data. The goal is to enable the model to learn meaningful representations by considering information from multiple images (or views) simultaneously. The model consists of a sequence of stages as shown in Figure \ref{fig:model}, each of which transforms the input data into increasingly abstract representations for downstream tasks such as classification, regression, or other types of predictions.

Let \(N\) denote the number of images, each of which has \(C\) channels, with spatial dimensions \(H\) (height) and \(W\) (width). The input to the model is a tensor of shape \( \mathbf{X} \in \mathbb{R}^{B \times (N \cdot C) \times H \times W} \), where \(B\) is the batch size. The input is processed in several stages: patch embedding, positional encoding, a Multi-view Attention Block, a transformer encoder, and finally a pooling and MLP head for prediction.

\textbf{Patch Embedding:} The first step in the MVVT model is to transform the input images into a set of fixed-size patches. Each image in the multi-view input is treated separately, and convolutional layers are applied to extract patch-level embeddings. Specifically, for each image \(i \in \{1, 2, \dots, N\}\), we extract the corresponding channels from the input tensor:
\[
\mathbf{X}_i = \mathbf{X}[:, (i*C):((i+1)*C), :, :]
\]
A convolution operation with a kernel of size \(P \times P\) and a stride \(P\) is then applied to each image to create patch embeddings:
\[
\mathbf{E}_i = \text{Conv2D}(\mathbf{X}_i, W_{\text{patch}}) \in \mathbb{R}^{B \times P_{\text{num}} \times D}
\]
where \(W_{\text{patch}} \in \mathbb{R}^{(C \times P \times P) \times D}\) is the learnable projection matrix, \(P_{\text{num}} = \frac{H}{P} \times \frac{W}{P}\) is the total number of patches per image, and \(D\) is the embedding dimension. 
This tensor represents the patch embeddings for the multi-view input, which will be used in the subsequent stages.

\textbf{Positional Encoding:} Vision Transformers, including the MVVT model, rely on positional encoding to preserve spatial relationships between image patches. Since the transformer architecture itself does not explicitly model spatial relationships, we add a learnable positional encoding to the patch embeddings. The positional encoding \( \mathbf{P} \in \mathbb{R}^{1 \times P_{\text{num}} \times (N \cdot D)} \) is added to the patch embeddings to retain positional information about the patches:
\[
\mathbf{Z}_i = \mathbf{E}_i + \mathbf{P}
\]

Here, \(\mathbf{Z}_i \in \mathbb{R}^{B \times P_{\text{num}} \times (N \cdot D)}\) is the tensor containing the patch embeddings with positional encodings.

\textbf{Multi-view Attention Block:} After applying positional encoding, we introduce a crucial component in the MVVT model: the Multi-view Attention Block (MAB). This block is designed to model interactions between patches from different images (views). The Multi-view Attention Block operates as follows:

1.\textbf{ Multi-Head Self-Attention (MSA):} The first operation in the MAB is the multi-head self-attention mechanism. This mechanism allows the model to learn relationships between patches within each image, as well as between patches from different images (i.e., between views). The multi-head self-attention is computed as:
\[
{\mathbf{Z}_i}' = \text{MSA}(\text{LN}(\mathbf{Z}_i)) + \mathbf{Z}_i
\]
where \(\text{LN}\) denotes layer normalization, and the self-attention mechanism aggregates information across all patches (both spatial and across images).

2. \textbf{Feedforward Network (MLP):} After applying the multi-head self-attention mechanism, the output \(\mathbf{Z}'\) is passed through a feedforward network (MLP), which refines the representations further. This is followed by another residual connection:
\[
\mathbf{Z}_i = \text{MLP}(\text{LN}(\mathbf{Z}_i')) + \mathbf{Z}_i'
\]
This step allows the model to transform the patch representations using nonlinearities, further enhancing the ability to capture complex relationships between the patches across multiple images.


This process is repeated for \(L\) layers, allowing the model to progressively learn more abstract and complex representations of the input data.

\textbf{Global Pooling and MLP Head:} After the transformer encoder, we apply mean pooling across all patch tokens to obtain a fixed-size representation of the multi-view input. Specifically, we perform the following mean pooling operation:
\[
\mathbf{Z}_{\text{pool}} = \frac{1}{P_{\text{num}}} \sum_{i=1}^{P_{\text{num}}} \mathbf{Z}_i
\]
This pooling operation aggregates information across all patches for each image and across all images in the multi-view input, resulting in a compact feature vector that encapsulates the information from the entire multi-view input.

The pooled feature vector \( \mathbf{Z}_{\text{pool}} \in \mathbb{R}^{B \times D} \) is then passed through an MLP head for the final task, which can be either classification or regression. The MLP head consists of fully connected layers with ReLU activations:
\[
\hat{y} = f_{\text{MLP}}(\mathbf{Z}_{\text{pool}})
\]
Here, \(f_{\text{MLP}}\) represents the fully connected layers, which can vary depending on the specific downstream task.


\section*{Experimental analysis and results}
The paper implemented a transformer-based baseline model to estimate plant age and leaf count. The experimental setup, performance metrics, and the corresponding results are discussed below.

\subsection*{Experimental setup}
The proposed MVVT model is trained separately on each of the four crops. To ensure generalizability, one plant from each crop is reserved for testing. The training and validation datasets are split in an 80:20 ratio from the training set. The model is trained using a batch size of 8, with 4 images per level, and optimized using Adam. Each input consists of 24 RGB images, resulting in 72 input channels since each image has 3 channels. The images are divided into patches of size 16 × 16 pixels, leading to 14 × 14 patches for a 224 × 224 image. Each patch is embedded into a 256-dimensional vector. The transformer model consists of 6 layers with 8 attention heads. The MLP head has a hidden dimension of 512. The model predicts a single value, either the day or leaf count, with a dropout rate of 0.1 applied to prevent overfitting.

\subsection*{Evaluation Metrics}

To assess the performance of the model, we use the following evaluation metrics: Root Mean Squared Error (RMSE) and Mean Absolute Error (MAE). The model is evaluated separately for plant age and leaf count.  

\subsubsection*{\textbf{Root Mean Squared Error (RMSE)}}
Root Mean Squared Error (RMSE) measures the average difference between the predicted and actual values. A lower RMSE indicates better model performance. It is defined as:

\begin{equation}
    RMSE = \sqrt{\frac{1}{N} \sum_{i=1}^{N} (y_i - \hat{y}_i)^2}
\end{equation}

where \( N \) is the number of samples, \( y_i \) is the actual value, and \( \hat{y}_i \) is the predicted value.

\subsubsection*{\textbf{Mean Absolute Error (MAE)}}
Mean Absolute Error (MAE) measures the average magnitude of errors between the predicted and actual values. Unlike RMSE, MAE does not square the errors, making it less sensitive to large errors. It is defined as:

\begin{equation}
    MAE = \frac{1}{N} \sum_{i=1}^{N} |y_i - \hat{y}_i|
\end{equation}

where \( N \) is the number of samples, \( y_i \) is the actual value, and \( \hat{y}_i \) is the predicted value. 

Both RMSE and MAE are computed separately for plant age and leaf count to analyze the model’s performance in each aspect.


\subsection*{Results}

Table \ref{tab:result} shows the performance of the MVVT model for two tasks: plant age prediction and leaf counting, evaluated separately for four different crops—Mustard, Radish, Okra, and Wheat. The results are measured using Root Mean Squared Error (RMSE) and Mean Absolute Error (MAE), where lower values indicate better performance.

For plant age prediction, the Radish crop performs best, achieving the lowest RMSE (7.31) and MAE (5.71). On the other hand, Mustard has the highest RMSE (13.18) and MAE (10.62), indicating the most challenging predictions.

For leaf counting, the Okra crop shows the best performance, with the lowest RMSE (2.27) and MAE (2.04). In contrast, Wheat has the highest errors, with RMSE (14.8) and MAE (10.8), making it the most difficult crop for leaf count estimation.

The average performance across all crops is 10.18 RMSE and 7.74 MAE for plant age prediction, and 6.89 RMSE and 5.52 MAE for leaf counting. These values summarize the model's overall performance across different plant types.


\begin{table}[ht!]
\centering
\begin{tabular}{lllll}
\hline
\multicolumn{1}{c}{\multirow{2}{*}{\textbf{Dataset}}} & \multicolumn{2}{l}{Age prediction} & \multicolumn{2}{l}{Leaf count} \\  
\multicolumn{1}{c}{} & \multicolumn{1}{l}{RMSE} & MAE & \multicolumn{1}{l}{RMSE} & MAE \\ \hline
Mustard & \multicolumn{1}{l}{13.18} & 10.62  & \multicolumn{1}{l}{5.95} & 4.91 \\ 
Radish & \multicolumn{1}{l}{\textbf{7.31}} & \textbf{5.71} & \multicolumn{1}{l}{4.90} & 4.34 \\ 
Okra & \multicolumn{1}{l}{8.03} & 5.86 & \multicolumn{1}{l}{\textbf{2.27}} &\textbf{ 2.04}  \\ 
Wheat & \multicolumn{1}{l}{11.6} & 8.8 & \multicolumn{1}{l}{14.8} &10.8  \\ \hline
Average & \multicolumn{1}{l}{10.18} & 7.74 & \multicolumn{1}{l}{6.89} &5.52  \\ \hline \\
\end{tabular}
\caption{Performance evaluation of plant age prediction and leaf count.}
\label{tab:result}
\end{table}
\section*{Conclusion}

This paper presents the GroMo challenge for two tasks: plant age prediction and leaf count estimation for mustard, okra, radish, and wheat using our proposed MVVT model. The performance of MVVT is evaluated using RMSE and MAE for each crop. The performance varies across crops due to differences in the number of plant instances, growth duration, and plant structure. For example, wheat performs the worst in leaf counting due to its growth style and longer growing period. Similarly, mustard performs the worst in age prediction because multiple plants exist within a single instance. In future work, we aim to optimize the model for better accuracy across different crops.

\balance

\bibliographystyle{ACM-Reference-Format}

\end{document}